\documentclass{svproc}

\usepackage[utf8]{inputenc}
\usepackage{graphicx}
\usepackage{hyperref}
\usepackage{multicol}
\usepackage{float}
\usepackage{listings}
\usepackage{amsmath}
\usepackage{amssymb}
\usepackage{eucal}
\usepackage{amsfonts}
\usepackage{footmisc}
\usepackage[]{algorithm2e}
\usepackage{amsfonts}

\usepackage{enumitem}

\title{Many Objective Bayesian Optimization}
\author{Lucia Asencio Mart\'in, Eduardo C. Garrido-Merch\'an}
\institute{Universidad Aut\'onoma de Madrid, Madrid, Spain\\
\email{lucia.asencio@estudiante.uam.es} \\ 
\email{eduardo.garrido@uam.es}}

\begin{document}

\maketitle

\begin{abstract}
Some real problems require the evaluation of expensive and noisy objective functions. Moreover, the analytical expression of these objective functions, and hence its gradients, may be unknown. These functions that are known as black-boxes, for example, estimating the generalization error of a machine learning algorithm and computing its prediction time in terms of its hyper-parameters. Multi-objective Bayesian optimization (MOBO) is a set of methods that has been successfully applied for the simultaneous optimization of black-boxes. Concretely, BO methods rely on a probabilistic model of the objective functions, typically a Gaussian process (GP). This model generates a predictive distribution of the objectives, capturing the uncertainty about their potential values. However, MOBO methods have problems when the number of objectives in a multi-objective optimization problem are $3$ or more, which is the many objective setting. In particular, the BO process is more costly as more objectives are considered, computing the quality of the solution via the hyper-volume is also more costly and, most importantly, we have to evaluate every objective function, wasting expensive computational, economic or other resources. However, as more objectives are involved in the optimization problem, it is highly probable that some of them are redundant and not add information about the problem solution. A measure that represents how similar are GP predictive distributions is proposed. We also propose a many objective Bayesian optimization algorithm that uses this metric to determine whether two objectives are redundant. The algorithm stops evaluating one of them if the similarity is found, saving resources and not hurting the performance of the multi-objective BO algorithm. We show empirical evidence in a set of toy, synthetic, benchmark and real experiments that GPs predictive distributions of the effectiveness of the metric and the algorithm.
\end{abstract}
\section{Introduction}
Optimization is the field that is concerned with finding the global extremum of an objective function in some region of interest. Assuming minimization, this scenario can be represented as: $\mathbf{x}^\star = \arg\min_{\mathbf{x} \in \mathcal{X}}f(\mathbf{x})$, where $f: \mathbb{R}^d \to \mathbb{R}$ is an example of the defined function, $\mathcal{X}$ is the region of interest or input space and $\mathbf{x}^\star$ is the global extremum belonging to the input space, which is a d-dimensional real-valued space $\mathbb{R}^d$. However, real problems include scenarios where the analytical expression of the function to be optimized is unknown. As a consequence, gradients are not accesible. For example, consider a scenario where the taste of a low calorie cookie needs to be optimized. There is no analytical expression for its taste. However, it is possible to perform evaluations by letting people taste the cookie to give their opinion about it. The objective functions that are expensive to evaluate, whose evaluations are contaminated by noise and that have no analytical expression, and hence no gradient information available, are called \textit{black-boxes}. For example, the estimation of the generalization error of ML algorithms is considered to be a black-box function. We find other applications in structure learning of probabilistic graphical models \cite{cordoba2018bayesian}, astrophysics \cite{balazs2021comparison} or even subjective tasks as suggesting better recipes \cite{garrido2018suggesting}.

Bayesian optimization (BO) is the class of methods that deal with the optimization of black-box functions with state-of-the-art results \cite{snoek2012practical}. In order to do so, BO uses a probabilistic surrogate model, typically a Gaussian process (GP), of the objective function. When not only one but several black-boxes are optimized, we deal with the Multi-objective BO scenario \cite{hernandez2016predictive}. If these objectives need to be optimized under the presence of constraints, we deal with the constrained multi-objective scenario \cite{garrido2019predictive,fernandez2020improved}. BO suggest one point per evaluation, but it also can suggest several points in the constrained multi-objective scenario \cite{garrido2020parallel}. Most critically, these problems involve the optimization of less than $4$ objectives. Many objective optimization has dealt with the optimization of more than $4$ objectives \cite{fleming2005many}. Interestingly, and as far as we know, this scenario has not been targeted by BO. A popular approach to solve the many objective optimization scenario is to get rid of objectives that can be explained through the other objectives. But in BO, objective functions are modelled via GPs. If we had a similarity measure of the predictive distribution $p(f(\mathbf{x})|\mathcal{D})$, where $\mathcal{D}$ is a dataset of previous observations, of the evaluation $y = f(\mathbf{x}) + \epsilon$ with $\epsilon$ being noise, of the black-box $f(\mathbf{x})$ computed by a GP over an input space $\mathcal{X}$, we could use it to propose an approach for the many objective BO scenario. This is precisely the motivation for this work: proposing a specialist GP predictive distribution similarity metric and use it in a many objective BO algorithm.

In particular, the purpose of our measure is to detect a GP that is so similar to another GP that we can stop fitting it in many objective BO. Hence, we cannot just apply the KL divergence, as it is just a similarity measure of probability distributions. Let $\boldsymbol{\theta}$ represent the set of parameters of a distribution $\mathcal{P}$. The KL divergence between two probability density functions $p(\boldsymbol{\theta})$ and $q(\boldsymbol{\theta})$ over continuous variables is given by the following expression:
\begin{equation}
\text{KL}(p(\boldsymbol{\theta})||q(\boldsymbol{\theta})) = \int_{-\infty}^{\infty} p(\boldsymbol{\theta})\text{log}(\frac{p(\boldsymbol{\theta})}{q(\boldsymbol{\theta})})d\boldsymbol{\theta}\,.
\end{equation}
As we can see, KL is not focused on things like the importance of the point and its neighbourhood that maximizes the objective function $f(\mathbf{x})$ but in all the probability distribution support. Our measure differs from KL divergence in the fact that we focus on particular characteristics of the GP predictive distribution that are relevant for discarding a GP in a many objective BO scenario.

In this work, we focus on comparing GP predictive distributions, as it is the arguably most widely used model in BO \cite{snoek2012practical}. Nevertheless, our measure could also be applied to the predictive distributions of Bayesian neural networks or Random forests, widening its applicability. After the description of the similarity metric, we propose a many objective Bayesian algorithm that uses this metric to delete redundant objective functions. Most interestingly, we show in a set of experiments that the performance given by the final recommendation of the BO algorithm does not suffer from the deletion of redundant GPs. Hence, we can obtain a solution for a many objective problem without losing a significant amount of quality and spending less resources than if we evaluate all the objective functions until the end of the BO process.

The paper is organized as follows. First, we give a detailed description of the fundamental concepts of Bayesian optimization. In particular, we describe the vanilla BO algorithm, Gaussian processes and multi-objective Bayesian optimization. These concepts will be useful to better understand the purpose of our proposed measure. Then, we include a section describing our proposed similarity measure. Once that all these concepts are understood, we illustrate the details of our main proposal, the many objective Bayesian optimization algorithm that reduces the redundant objectives of a many objective Bayesian optimization setting. We add empirical evidence of the practical use of this measure and this algorithm in an experiments section concerning toy, synthetic, benchmark and real problems to support the claim that our algorithm is useful in practice and effectively identifies redundant objectives without affecting the performance of BO significantly. Lastly, we illustrate a series of conclusions about this work and further lines of research. 
\section{Fundamentals of Bayesian Optimization}
In this section, we will formally explain the fundamentals of Bayesian optimization in detail. In particular, BO is an iterative algorithm that seeks to retrieve the extremum $\mathbf{x}^\star$ of a black-box function $f(\mathbf{x})$ where $\mathbf{x} \in \mathcal{X}$ and $\mathcal{X}$ is the input space where $f(\mathbf{x})$ can be evaluated.
\begin{equation}
\mathbf{x}^\star = \arg\min_{\mathbf{x} \in \mathcal{X}}f(\mathbf{x})\,,
\end{equation}
assuming minimization. In order to do so, BO places a probabilistic surrogate model (for example a GP) over $\mathcal{X}$ that computes a predictive distribution of the evaluation of $\mathbf{x}$, $p(f(\mathbf{x})|\mathcal{D})$, where $\mathcal{D} = \{(\mathbf{x}_i, y_i)| i = 1,...,t\}$ is the dataset of previous observations at iteration $t$. BO computes an acquisition function $\alpha(\mathbf{x}; p(f(\mathbf{x})|\mathcal{D}))$ that computes the expected utility of evaluating $\mathbf{x}$ to retrieve the extremum $\mathbf{x}^\star$. An acquisition function $\alpha(\mathbf{x})$ represents an exploration-exploitation trade-off. Specifically, the acquisition function $\alpha(\mathbf{x})$ favors exploration in the sense that it is high in areas where no point $\mathbf{x}$ has been evaluated before. These areas may contain values near the optimum value. It also needs to perform exploitation, \textit{i.e.}, the acquisition function $\alpha(\mathbf{x})$ favors the evaluation of points that are near of good evaluations. Concretely, we expect that the new evaluations are near the optimum as we assume that the function is smooth. The point suggested for evaluation at every iteration $t$ of the BO algorithm is the one that maximizes the acquisition function.
\begin{equation}
\mathbf{x}_t  = \arg\max_{\mathbf{x} \in \mathcal{X}}\alpha_t(\mathbf{x})\,.
\end{equation}
The trick of BO is that optimizing the acquisition function $\alpha(\mathbf{x})$ is a cheap process as computing $\alpha(\mathbf{x})$ depends on the information given by the predictive distribution $p(f(\mathbf{x})|\mathcal{D})$, not on the evaluations of the objective function $f(\mathbf{x})$. In particular, the acquisition function $\alpha(\mathbf{x})$ will only have as many dimensions $D$ as dimensions has the black-box being optimized $f(\mathbf{x})$. We can compute the gradients of the acquisition function and optimize it via a grid search and an optimizer such as L-BFGS. The steps of BO are summarized on Algorithm \ref{alg:bo}. Details for computing the predictive distribution $p(f(\mathbf{x})|\mathcal{D})$ that is used to build the acquisition function $\alpha(\mathbf{x})$ at every iteration $t$ of the BO algorithm are explained in the following subsection, along with specific details of the probabilistic surrogate model used in this work, the Gaussian process.
\begin{figure}[tb]
\begin{algorithm}[H]
\label{alg:bo}
\textbf{Input:} Maximum number of evaluations $T$.
\caption{BO of a black-box objective function $f(\mathbf{x})$.}
\For{$\text{t}=1,2,3,\ldots,T$}{
        {\bf 1:} {\bf if $N=1$:}\\
        \hspace{.7cm}Choose $\mathbf{x}_t$ at random from $\mathcal{X}$. \\
        \hspace{.5cm}{\bf else:} \\
        \hspace{.7cm}Find $\mathbf{x}_t$ by maximizing the acquisition function:
        $\mathbf{x}_t = \underset{\mathbf{x} \in \mathcal{X}}{\text{arg max}} \quad \alpha_t(\mathbf{x})$.

        {\bf 2:} Evaluate the black-box objective $f(\cdot)$ at $\mathbf{x}_t$: $y_t=f(\mathbf{x}_\text{t}) + \epsilon_t$.

        {\bf 3:} Augment the dataset with the new observation: $\mathcal{D}_{1:t}=\mathcal{D}_{1:t-1} \bigcup \{\mathbf{x}_t, y_t\}$.

        {\bf 4:} Fit again the GP model using the augmented dataset $\mathcal{D}_{1:t}$.
 }
{\bf 5:} Obtain the recommendation $\mathbf{x}^\star$: Point associated with the value that optimizes the GP prediction or with the best observed value. \\
\KwResult{Recommended point $\mathbf{x}^\star$}
\vspace{.5cm}
\end{algorithm}
\end{figure}
\subsection{Gaussian Processes}
\label{}
A Gaussian Process (GP) is a collection of random variables (of potentially infinite size), any finite number of which have (consistent) joint Gaussian distributions \cite{rasmussen2003gaussian}. We can also think of GPs as defining a distribution over functions where inference takes place directly in the space of functions \cite{rasmussen2003gaussian}. A GP can be used for regression of a function $f(\mathbf{x})$.

Let $\mathbf{X}=(\mathbf{x}_1,...\mathbf{x}_N)^T$ be the training matrix and $\mathbf{y}=(y_1,...,y_N)^T$ be a vector of labels to predict. We define as a dataset $\mathcal{D} = \{(\mathbf{x}_i, y_i)| i = 1,...,n\}$ the set of labeled instances. A GP is fully characterized by a zero mean and a covariance function $k(\mathbf{x},\mathbf{x}')$, that is, $f(\mathbf{x}) \sim \mathcal{G}\mathcal{P}(\mathbf{0},k(\mathbf{x},\mathbf{x}'))$. 

Given a set of observed data $\mathcal{D} = \{(\mathbf{x}_i, y_i)| i = 1,...,n\}$, where $y_i=f(\mathbf{x}_i) + \epsilon_i$ with $\epsilon_i$ some
additive Gaussian noise, a GP builds a predictive distribution for the potential values of $f(\mathbf{x})$ at a new input point $\mathbf{x}^\star$. This distribution is Gaussian. The GP mean, $\mu(\boldsymbol{x})$, is usually set to $0$. Namely,
$p(f(\mathbf{x}^\star)|\mathbf{y}) =\mathcal{N}(f(\mathbf{x}^\star)|
\mu(\mathbf{x}^\star),  v(\mathbf{x}^\star))$, where the mean $\mu(\mathbf{x}^\star)$ and variance $v(\mathbf{x}^\star)$ are respectively given by:
\begin{align}
\mu(\mathbf{x}^\star) & = \mathbf{k}_{\star}^{T} (\mathbf{K}+\sigma^{2}\mathbf{I})^{-1}\mathbf{y}\,, \\
v(\mathbf{x}^\star) & = k(\mathbf{x}_{\star},\mathbf{x}_{\star}) - \mathbf{k}_{\star}^T(\mathbf{K}+\sigma^{2} \mathbf{I})^{-1}\mathbf{k}_\star\,,
\label{eq:pred_dist}
\end{align}
where $\mathbf{y}=(y_1,\ldots,y_N)^\text{T}$ is a vector with the observations collected so far;
$\sigma^2$ is the variance of the additive Gaussian noise $\epsilon_i$;
$\mathbf{k}_\star = \mathbf{k}(\mathbf{x}_*)$ is a $N$-dimensional vector with the prior covariances between the test point $f(\mathbf{x}^\star)$ and
each of the training points $f(\mathbf{x}_i)$; and $\mathbf{K}$ is a $N\times N$ matrix with the prior covariances
among each $f(\mathbf{x}_i)$, for $i=1,\ldots,N$. Each element $\mathbf{K}_ij = k(\mathbf{x}_i, \mathbf{x}_j)$ of the matrix $\mathbf{K}$ is given by the covariance function between each of the training points $\mathbf{x}_i$ and $\mathbf{x}_j$ where $i,j = 1,...,N$ and $N$ is the total number of training points. The particular characteristics assumed for $f(\mathbf{x})$ (\emph{e.g.}, level of smoothness, additive noise, etc.) are specified by the covariance function $k(\mathbf{x},\mathbf{x}')$ of the GP. A popular example of covariance function is the squared exponential, given by:
\begin{align}
k(\mathbf{x},\mathbf{x}') & = \sigma^2_f \exp\left(-\frac{r^2}{2\ell^2}\right) + \sigma^2_n\delta_{pq}\,,
\end{align}
where $r$ is the Euclidean distance between $\mathbf{x}$ and $\mathbf{x}'$, $\ell$ is a hyper-parameter known as length-scale that controls the smoothness of the functions generated from the GP, $\sigma^2_f$ is the amplitude parameter or signal variance that controls the range of variability of the GP samples and $\sigma^2_n\delta_{pq}$ is the noise variance that applies when the covariance function is computed to the same point $k(\mathbf{x},\mathbf{x})$. Those are hyper-parameters of the GP. Let $\boldsymbol{\theta}$ be the set of all those hyper-parameters. We can find point estimates for the hyper-parameters $\boldsymbol{\theta}$ of the GP via optimizing the log marginal likelihood. The marginal likelihood is given by the following expression:
\begin{equation}
\log p(\mathbf{y}|\mathcal{X}, \boldsymbol{\theta}) = - \frac{1}{2} \mathbf{y}^T (\mathbf{K} + \sigma_n^{2} I)^{-1} \mathbf{y} - \frac{1}{2} \log | \mathbf{K} + \sigma_n^2I| - \frac{n}{2} \log 2\pi\,.
\end{equation}
The previous analytical expression can be optimized to obtain a point estimate $\boldsymbol{\theta}^\star$ for the hyper-parameters $\boldsymbol{\theta}$. We can optimize it through a local optimizer such as L-BFGS-B \cite{zhu1997algorithm} and via the analytical expression of the marginal likelihood gradient $\nabla_{\boldsymbol{\theta}} \log (\mathbf{y}|\mathbf{X}, \boldsymbol{\theta}) = (\partial \log (\mathbf{y}|\mathbf{X}, \boldsymbol{\theta} /\partial \theta_1 , ..., \partial \log (\mathbf{y}|\mathbf{X}, \boldsymbol{\theta}) / \partial \theta_M )^T$ whose partial derivatives are given by:
\begin{align}
\frac{\partial}{\partial \theta_j} \log (\mathbf{y}|\mathbf{X}, \boldsymbol{\theta}) & = \frac{1}{2} \mathbf{y}^T \mathbf{K}^{-1} \frac{\partial \mathbf{K}}{\partial \theta_j} \mathbf{K}^{-1} \mathbf{y} - \frac{1}{2}\text{tr}(\mathbf{K}^{-1}\frac{\partial \mathbf{K}}{\partial \theta_j}) \nonumber \\
& = \frac{1}{2} \text{tr} ((\boldsymbol{\alpha}\boldsymbol{\alpha}^{T} - \mathbf{K}^{-1})\frac{\partial \mathbf{K}}{\partial \theta_j}) \,,
\end{align}
where $\boldsymbol{\alpha} = \boldsymbol{K}^{-1}\mathbf{y}$ and $M$ is the number of hyper-parameters.
\subsection{Multi-objective Bayesian Optimization and Many Objective Optimization}
Until this subsection, we have explored the BO algorithm applied to obtaining the optimum $\mathbf{x}^\star$ of a single objective function $f(\mathbf{x})$ modelled by a GP. However, we can consider the simultaneous optimization of a set of $k$ independent objective functions $\mathbf{f}(\mathbf{x}) = f_1(\mathbf{x}),...,f_K(\mathbf{x})$. Each of these functions can be modelled by a GP. In the multi-objective setting, it is not possible to optimize all the objective functions $\mathbf{f}(\mathbf{x})$ simultaneously, as they may be conflicting. In spite of this, it is still possible to
find a set of optimal points $\mathcal{X}*$  known as the Pareto set \cite{siarry2003multiobjective}. More formally, we define that the point $\mathbf{x}$
dominates the point $\mathbf{x}'$ if $f_k (\mathbf{x})\leq f_k (\mathbf{x}')$ $\forall k$,
with at least one inequality being strict. The Pareto set is considered to be optimal because for each point in that set one cannot improve in one of the objectives without deteriorating some other objective. Given $\mathcal{X}^*$, a final user may then choose a point from this set according to their preferences.

When we optimize a single objective function $f(\mathbf{x})$, we can evaluate its solution $\mathbf{x}$ by minimizing its regret $r$ with respect to the optimum of the problem $\mathbf{x}^\star$. Recall that the regret can be computed as: $r = |f(\mathbf{x}^\star) - f(\mathbf{x})|$. This metric is not applicable to the multi-objective scenario. In this case, we need to define a different metric to evaluate the quality of an estimate of the Pareto set $\hat{\mathcal{X}}^\star$. The metric that is usually used to evaluate the quality of an estimated Pareto set $\hat{\mathcal{X}}^\star$ is the hypervolume. The hypervolume is the area covered by the Pareto frontier $\hat{\mathcal{Y}}^\star$ with respect to a reference point in the space. There exist several mechanisms to efficiently compute the hypervolume metric \cite{while2006faster}.

However, computing the hypervolume for more than $3$ objectives is an expensive process. Hence, most multi-objective BO settings include $3$ or less objectives. In the optimization literature, when more than $3$ objectives are included in the optimization problem we refer to this scenario as the Many Objective scenario. In particular, a popular strategy of this scenario is to get rid of objectives that are correlated and, hence, do not add significant information about the location of the Pareto set in the input space as their functional form is very similar as one of the other objectives. Hence, in the Bayesian optimization setting, a strategy to solve the Many Objective scenario would be to detect those correlations and convert a many objective setting into a multi-objective setting, that is, an optimization scenario with $3$ or less objectives. As we model each of the objective functions with a Gaussian process, to propose such an algorithm, we first need a similarity measure between Gaussian processes, that would represent how similar are their predictive distributions. 

\section{The Similarity Measure}
\label{sec:similarity_measure}
%\textbf{INSPIRATION Frases/palabras que parecen utiles y quiero tener a mano:}
%\begin{itemize}
%    \item We give the details... We present... 
%    \item For simplicity, the expression given... comes from... as in...
%    \item Proposed, more sofisticated, computing
%    \item Note that... Let...
%\end{itemize}
Let $f(\textbf{x})$, $g(\textbf{x})$ be two GPs. Let us work under the assumption that their covariance functions $k(\mathbf{x}, \mathbf{x})$ have similar analytical expressions (e.g., both are squared exponential functions). Given a set $\mathbf{X}_{\star}$ of input points, we can compute $\mu_f(\textbf{X}_{\star})$ and $\mu_g(\textbf{X}_{\star})$, the predicted mean vectors for each process, as well as $v_f(\textbf{X}_{\star}(\textbf{X}_{\star})$ and $v_g(\textbf{X}_{\star})$, their covariance matrices.\\
We now define a notion of distance between these two processes, firstly  presenting its mathematical expression:

\begin{align} 
d(f(\textbf{x}) , g(\textbf{x})) =&  \varepsilon_1d_1\left(T(\mu_f(\textbf{X}_{\star})), \mu_g(\textbf{X}_{\star}), \delta\right) + \nonumber \\ 
&  \varepsilon_2d_2\left(v_f(\textbf{X}_{\star}), v_g(\textbf{X}_{\star})\right) + \nonumber \\
&  \left( 1 - \varepsilon_1 - \varepsilon_2\right)\left(1 - \rho\left(\mu_f(\textbf{X}_{\star}), \mu_g(\textbf{X}_{\star})\right)\right) \,.
\end{align}
%$$d(f(\textbf{x}) , g(\textbf{x})) = \varepsilon_1d_1\left(T(\mu_f(\textbf{X}_{\star})), \mu_g(\textbf{X}_{\star}), \delta\right) + \varepsilon_2d_2\left(v_f(\textbf{X}_{\star}), v_g(\textbf{X}_{\star})\right) + \left( 1 - \varepsilon_1 - \varepsilon_2\right)\left(1 - \rho\left(\mu_f(\textbf{X}_{\star}), \mu_g(\textbf{X}_{\star})\right)\right)$$
As it can be seen, the measure is given following a weighted sum model (WSM \cite{TriantaphyllouEvangelos2000Mdmm}). The WSM contains three components to which we will refer as
\begin{align}
s_1 = \varepsilon_1d_1\left(T(\mu_f(\textbf{X}_{\star})), \mu_g(\textbf{X}_{\star}) \delta\right) \,, \nonumber \\
s_2 = \varepsilon_2d_2\left(v_f(\textbf{X}_{\star}), v_g(\textbf{X}_{\star})\right) \,, \nonumber \\
s_3 = \left( 1 - \varepsilon_1 - \varepsilon_2\right)\left(1 - \rho\left(\mu_f(\textbf{X}_{\star}), \mu_g(\textbf{X}_{\star})\right)\right) \,.
\end{align}
The objective of $s_1$ and $s_3$ is to describe the distance between both mean vectors, while $s_2$ aims to reflect the distance between the covariance matrices. We will now analyze each of these components and their respective parameters. The first one, $s_1$, is given in terms of a tolerance $\delta$, a transformation function $T$ and a distance function $d_1$ between the mean vectors. The election of $T$ describes when two  mean vectors $\mu_f(\textbf{X}_{\star}) \neq \mu_g(\textbf{X}_{\star})$ should be considered equal. For example, since we will be optimizing these vectors, if $\mu_f(\textbf{X}_{\star}) = 2\mu_g(\textbf{X}_{\star})$, their critical points will be the exact same and we might want to consider them as a single vector. Although $T$ can be chosen by the user depending on their needs, the proposed implementation provides a function $T(\mu_f(\textbf{X}_{\star})) = a\mu_f(\textbf{X}_{\star})+ b\textbf{1}$, where $a>0$ and $b$ are scalars chosen with the least squares method to give the best fit of  $\mu_f(\textbf{X}_{\star})$ onto $\mu_g(\textbf{X}_{\star})$. This transformation reflects the fact that two vectors that are proportional and whose difference is constant behave in the same way in terms of optimization.

With function $d_1(\cdot)$, the user is able to choose in which way they want to measure the distance between $T(\mu_f(\textbf{X}_{\star}))$ and $\mu_g(\textbf{X}_{\star})$. Several options are given to the user in our implementation, each of them being convenient depending on the nature of the problem modeled by the GP. Some of these options are to define $d_1$ as the number (or percentage) of points where $T(\mu_f(\textbf{X}_{\star}))\neq\mu_g(\textbf{X}_{\star}) $, or as a $p$-norm ($d_1= \|{\mu_g(\textbf{X}_{\star}) - T(\mu_f(\textbf{X}_{\star}))}\|_{p} $) which includes euclidean norm, infinity norm, etc.
Lastly, with $\delta$ the user is allowed to change the desired level of tolerance given to $d_1$, i.e. the distance is calculated only among the vectors' elements where the chosen $d_1$ is greater than $\delta$ in its element-wise operations. For example, if we chose a $1$-norm as $d_1$, $s_1$ would be computed as $\sum_{|\mu_g - T(\mu_f)| > \delta}{|\mu_g(\textbf{X}_{\star}) - T(\mu_f(\textbf{X}_{\star}))|}$.\\\\
The other weighted sum term used to compare the two mean vectors is $s_3$. It is the only fixed term in the sum, and it represents the Pearson correlation coefficient between the GP means. 
This coefficient is defined as 
\begin{equation}
{\rho(\mu_f(\textbf{X}_{\star}),\mu_g(\textbf{X}_{\star}))={\frac {\mathbb{E} [(\mu_f(\textbf{X}_{\star})-\overline{\mu _f(\textbf{X}_{\star})})(\mu_g(\textbf{X}_{\star})-\overline{\mu _y(\textbf{X}_{\star})}))]}{\sigma_{\mu_f}\sigma _{\mu_g}}}} \,,
\label{eq:pearson}
\end{equation}
where $\overline{\mu(\cdot)}$ denotes the mean value of a vector $\mu$ and $\sigma_\mu$ is its standard deviation.\\The reason we were first interested in this operator is because of its interpretation. The coefficient $\rho(\mu_f(\textbf{X}_{\star}), \mu_g(\textbf{X}_{\star}))$ ranges from -1 to 1. If it equals 1, there is a (positive) linear equation describing $\mu_g(\textbf{X}_{\star})$ in terms of $\mu_f(\textbf{X}_{\star})$; if it equals -1, this linear equation has a negative slope and, when it is 0, no linear correlation between $\mu_f(\textbf{X}_{\star})$ and $\mu_g(\textbf{X}_{\star})$ exists. 
Moreover, following Eq. (\ref{eq:pearson}), $\rho$ increases whenever $\mu_f(\textbf{X}_{\star})$ and $\mu_g(\textbf{X}_{\star})$ both increase or decrease. It decreases when their growth behaviour is different. This is very valuable for our problem, since we need to identify whether two vectors are increasing and decreasing in a similar fashion, i.e., their maximums and minimums lie around the same positions.\\
We found this to be the most accurate way of detecting similar processes, since it detects that sample vectors of functions like $x^6$ and $x^2$, which would a priori seem very different using any conventional vector distance (one grows much faster than the other) behave essentially the same: they both decrease from $-\infty$ to 0, have a minimum in 0 and increase towards $\infty$.
Lastly, the addend $s_2$ is intended to measure the distance between both predictive variances, and therefore any matrix norm could be used for this purpose.
We have not found the matrices distance to be significant when it comes to deciding whether two GPs should be optimized analogously, although this might be because of working under the assumption that $f(\textbf{X}_{\star})$ and $g(\textbf{X}_{\star})$ have similar covariance functions.  In  case the user wants to make use of the matrices similarity, note that entrywise matrix norms should be preferred over the ones induced by vector norms because of their lower computational cost \cite{0908.1397}.\\ 
As future work, these matrices could be used to measure the uncertainty of the distance between two GPs, since they represent the uncertainty of the GPs' predictions.
\section{Many Objective Bayesian Optimization algorithm}

As far as we know, Many objective Bayesian optimization has not been targeted by the literature before. Hence, as it is a problem that it is important to solve for real problems, we propose in this work a solution based on reducing the objectives of the optimization problem. Our purpose is to detect when many objective optimization problems, dealing with more than $3$ objectives, can be reduced to the multi-objective optimization setting, that deals with $3$ or less objectives. Concretely, when such a situation is detected, we get rid of redundant objectives and solve the new problem using a multi-objective Bayesian optimization method. 

Let us formalize our proposed algorithm for many objective Bayesian optimization. Recall that in each iteration of Bayesian optimization, the GPs that are modelling the objective functions $\mathbf{f}(\mathbf{x})$ are fitted to a set of observed data $\mathcal{D}$ and an acquisition function $\alpha(\mathbf{x})$ is built from the predictive distribution of the GPs. We compare the $k$ predictive distributions $p(f_k(\mathbf{x})| \mathcal{D}, \mathbf{x})$ of a GP with the similarity measure proposed in Section \ref{sec:similarity_measure} to detect when two GPs are similar to each other and, as we could determine the Pareto set by simply evaluating one of them, we could stop evaluating one of them to save computational or other resources. 

Let $f_1(\mathbf x),...,f_d(\mathbf x)$ be the objective functions. At iteration $n$, let $\mathbf X = (\mathbf x_1, ..., \mathbf x_n)^T$ be a matrix whose rows represent the observed points $\mathbf x_i$. For each $\mathbf x_i$, let $\mathbf y_i = (y_{i1},... y_{id})$ be a vector that represents the noisy evaluations of the objectives in the point  $\mathbf x_i$, that is, $y_{ij} = f_j(\mathbf x_i) + \epsilon_{ij}$ where $\epsilon_{ij} \sim \mathcal{N}(0, \sigma)$ for a configurable $\sigma$. Let $\mathbf Y = (\mathbf y_1,..., \mathbf y_n)^T$ be a matrix with all the evaluations and $\mathcal{D} = (\mathbf{X}, \mathbf{Y})$ the dataset with the observed points and its respective evaluations in all the black-boxes.

After computing each predictive distribution $p_k(f_k(\mathbf{x}|\mathcal{D})$ for every objective, we can compute the set of similarities between each of the black-boxes. Let $D$ be the set of those similarities, that is defined by $D = \{ d_{ij} = d(f_i(\mathbf x), f_j(\mathbf x) )\; | \; 1<i<j<n\}$ where $i$ and $j$ represent indexes of the objective functions. In particular, the size of the set of similarities would be $|D| = k!/(2!(k-2)!)$ where $k$ is the number of objectives. In every iteration, our algorithm computes the set of similarities $D$ and traverses it. If it finds a value $d_{ij}<\varepsilon$ for a certain predefined $\varepsilon$, our algorithm will delete the objective $f_i(\mathbf x)$ from the optimization as it turns out to be redundant, and hence, in the following iteration the optimization would run with one objective less.

Deleting one objective from the optimization problem involves losing information about the Pareto set $\mathcal{X}$ of the full optimization problem. Following the previous criterion, similar objectives are deleted, but as they may be not equal, the solution of the reduced multi-objective optimization problem is not identical to the full optimization problem. In other words, quality of the final recommendation will be lost. However, the practitioner may be interested in sacrificing quality and saving a huge amount of resources (such as computational time, money or avoiding contamination) due to not evaluating redundant objectives. By performing this algorithm, depending on the value assigned to the deletion threshold $\varepsilon$ we are dealing with a trade-off involving quality and resources. In order to ensure not losing too much information concerning the Pareto set $\mathcal{X}$ in a single iteration, our algorithm limits to $1$ the number of objectives that can be reduced from the problem in a single iteration. Moreover, at the beginning of the Bayesian optimization process, the uncertainty computed by the predictive distributions about the problem is too high, hence, although two objectives may be correlated, there is not enough information recovered by the objective functions to predict their similarity. Depending on prior knowledge about the problem, we may know if the functions are smooth, if there is a high probability that a lower dimensional manifold explains the variability of the objective functions or other specific features about them. Hence, we may have some confidence that, in certain problems, after evaluating the objectives a certain amount of times we may start computing the similarity set $D$ and, in others, we may have to wait until more evaluations are performed. As a consequence, we include a hyper-parameter for our algorithm, $\delta$, that represents the index of the iteration when the algorithm will start looking for redundant objectives until the end of the optimization problem. To sum up, Algorithm \ref{maobo} summarizes the steps of our many objective Bayesian optimization proposed method.

\begin{algorithm}
\label{maobo}
      \SetKwData{In}{In}
      %\SetKwFunction{RedundantObjective}{RedundantObjective}
      \SetKwInOut{Input}{Input}
      \SetKwInOut{Output}{Output}
      \SetKwData{Distances}{Distances}
      \SetKwData{Ndim}{Ndim}
      \SetKwFunction{Delete}{Delete}
      \SetKwFunction{Length}{Length}
      \SetKwFunction{Distance}{Distance}
      \SetKwFunction{Grid}{Grid}
      \Input{An integer $iter \geq 0$ representing the BO iteration, an integer $\delta\geq 0$ that allows to set the iteration number when the reduction begins to be performed, a real number $0\leq \varepsilon \leq 1$ representing a tolerance, $obj$ is a vector of length $dim$ that contains the objective functions $\mathbf{f}(\mathbf{x})$. }
      \Output{The list of possibly reduced objective functions $\mathbf{f}(\mathbf{x})$.}
      \BlankLine
      %\emph{special treatment of the first line}\;
      \If{$ iter \geq \delta$}{
            \Ndim $\leftarrow$ \Length{obj}\;
            \For{$i\leftarrow 1$ \KwTo \Ndim}{
                \For{$j\leftarrow i+1$ \KwTo \Ndim}{
                    \Distances$[obj[i],obj[j]] \leftarrow $ \Distance{$obj[i]$, $obj[j]$}\;
                }
            }
           \ForEach{\Distance$[i,j] \in $\Distances}{
                \If{\Distances$[i,j]< \varepsilon$ }{
                    \Return \Delete{$obj[i]$}
                }
            }
      }
\caption{Many objective Bayesian optimization algorithm based on the reduction of redundant objectives according to the similarity measure of GP predictive distributions.}
\end{algorithm}

\section{Related Work}
  In spite of the fact that in Bayesian optimization the many objective setting has not been targeted, several approaches have been proposed to deal with many objective optimization in the optimization literature. In particular, we can find that the many objective setting can be transformed into a single objective scenario \cite{yang14,bentley98}. However, this solution will only generate one point as a recommendation, and not a Pareto set of points where each one of them represent a different utility of every objective involved into the many objective setting. As a consequence, we find that if the objectives are conflicting the final solution will only represent a particular utility over the different objectives and we will lose particular solutions that are better for certain objectives at the cost of not being good for other ones. As the final utility of the user may be to give more importance to certain objectives or just to consider several utilities given by the points belonging to the Pareto set, we do not consider this solution as a proper one to the many objective optimization setting. We also find other alternatives to solve the many objective scenario that are not focused on the efficiency given by the Pareto set to compare two potential solutions of the problem \cite{garzafabre09}. Another approach, based on game theory, selects a single Pareto set option as the solution of the many objective scenario \cite{binois20}.
  
  Recently, a generalization of pigeon-inspired optimization, based on swarm intelligence and in the behaviour of carrier pigeons, has been proposed to tackle the design of a many objective optimization algorithm \cite{cui19}. However, metaheuristics such as pigeon-inspired optimization follow the assumption that the black-boxes that we evaluate are not very expensive. That is, we can afford a high number of evaluations of the optimization problem. As a consequence, we can not use approaches derived from metaheuristics in the Bayesian optimization scenario, where we assume that the black-boxes are expensive to evaluate and we can not afford the necessary number of iterations that an approach based on a metaheuristic needs to obtain a good solution for the many objective optimization scenario. Finally, the Pareto corner search evolutionary algorithm (PCSEA) has also been proposed. As in our proposal, the algorithm reduces the number of objectives by detecting which objectives are irrelevant and deleting them. However, our algorithm differs with respect to PCSEA in the fact that we are dealing with a Bayesian optimization algorithm so the way to detect if an objective is irrelevant is to compare the predictive distribution of the surrogate probabilistic model, in our case a GP, with the others \cite{singh2011pareto}. 
  
  Concerning applications, many objective algorithms have been applied to optimize the controller of a hybrid car \cite{narukawa12}. It has also been used for software refactorization \cite{mkaouer14}, for the optimization of the design of prefabricated industrial buildings \cite{bandyopadhyay14} and in planning nursing schedules \cite{otake10}. Finally, it has also been applied for significant scenarios in aeronautics \cite{wickramasinghe10,asafuddoula14}. 
  
  Most critically, despite the variety of solutions that exist for many objective optimization, this scenario has not been tackled for the Bayesian optimization setting, where the objectives are very expensive to evaluate and, as it has been said before, we can not rely on metaheuristic approaches. As a consequence, we believe that our approach makes a significant contribution for the Bayesian optimization community, as a first approach to tackle the Many objective scenario using a Bayesian optimization method. Concretely, based on reducing the dimension of the objective space, by the detection of similar predictive distributions \cite{asencio2021similarity}.

\section{Experiments}
We now illustrate the different experiments that we have performed throughout this project to add empirical evidence to support the claim that our method is useful and can be applied to real scenarios. We begin the section with an empirical study of the proposed similarity measure, and then, with a set of experiments showing how the many objective Bayesian algorithm effectively identifies redundant GPs and removes them from the optimization problem. 

\subsection{Testing the GP predictive distribution similarity measure}
\begin{figure}[t]
\includegraphics[width=\textwidth]{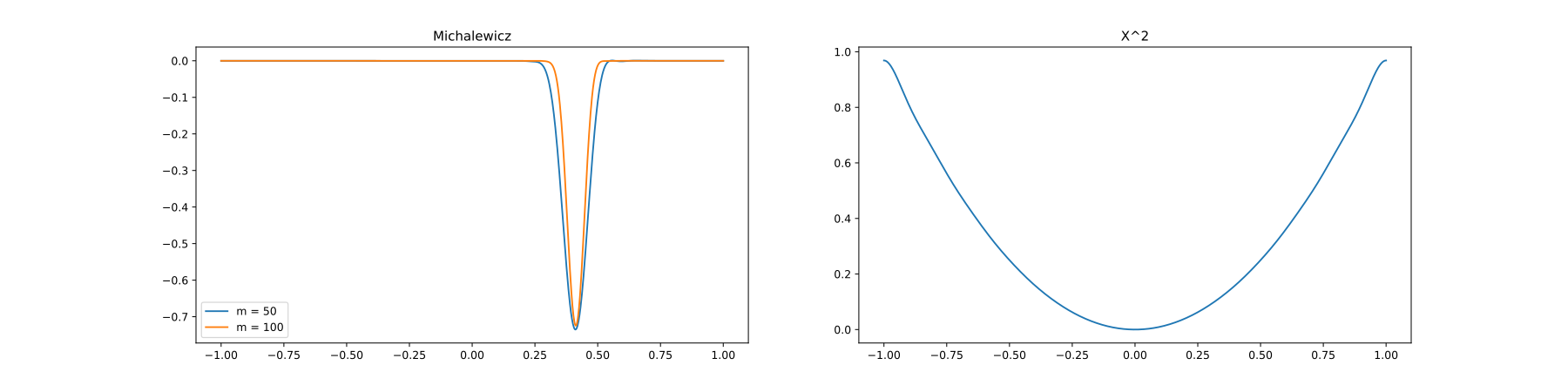}
\centering
\caption{Toy functions for one dimension}
\label{fig:1d}
\end{figure}

\begin{figure}[t]
\includegraphics[width=\textwidth]{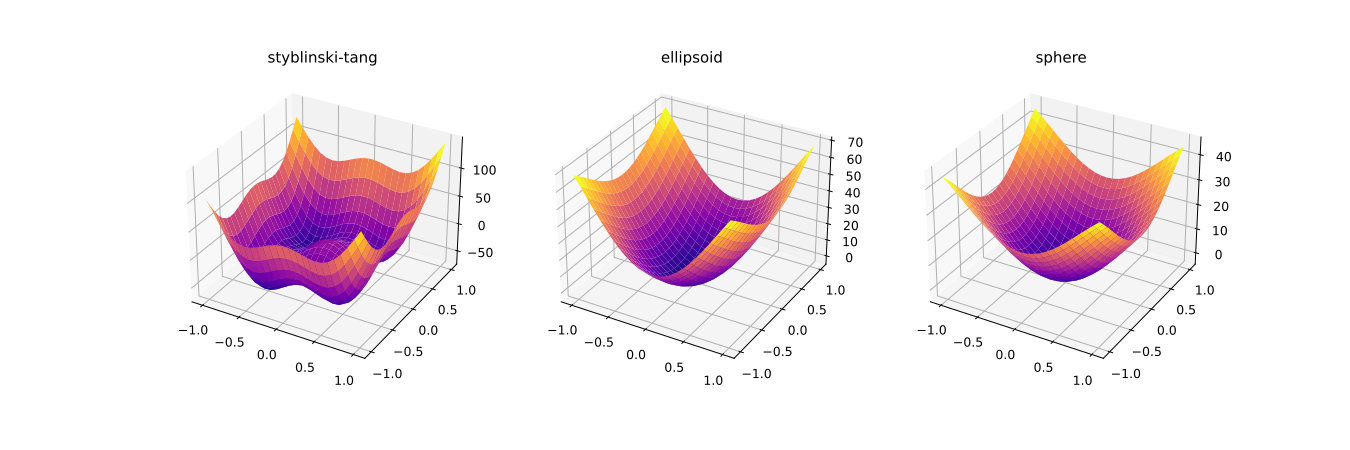}
\centering
\caption{Bowl-shaped functions}
\label{fig:bowl}
\end{figure}

\begin{figure}[t]
\includegraphics[width=\textwidth]{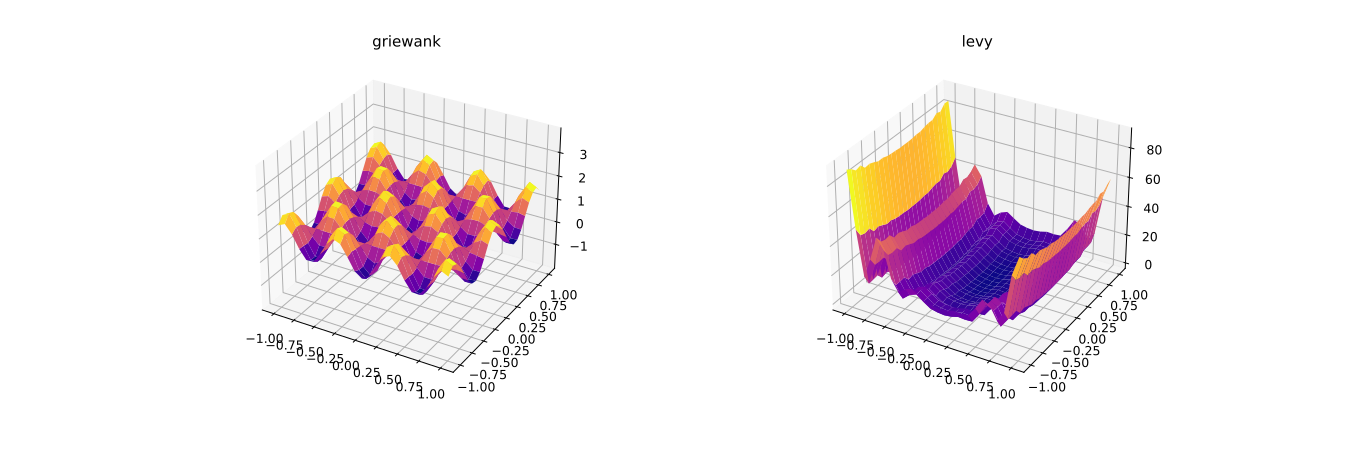}
\centering
\caption{Griewank and Levy functions}
\label{fig:misc}
\end{figure}

\begin{figure}[t]
\includegraphics[width=\textwidth]{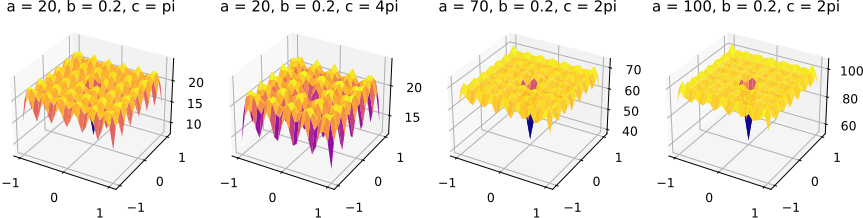}
\centering
\caption{Ackley functions}
\label{fig:ackley}
\end{figure}

For all the experiments, we have set the parameter $\delta=0$ and, since we didn't found the covariance matrices distance to be significant under our hypothesis, we used $\varepsilon_1 = 0.25$ and $\varepsilon_2 = 0$ (therefore $\rho$'s weight is $0.75$). For the transformation $T$, the previously explained linear transformation using the least squares method was used. We have chosen $d_1$ to be the average relative distance between the points in the two mean vectors, i.e., the mean of the vector given by $|T\left(\mu_f(\textbf{X}_\star)\right) - \mu_g(\textbf{X}_\star)|$ divided by the subtraction of the greatest element found in the two vectors minus the smallest.  \\For the following examples, various GPs were fitted taking sample vectors from benchmark functions. We will now discuss some of the results obtained by applying our measure to find the distance between them.\\\\
We will start with some uni-dimensional toy functions. We have chosen to compare three GPs from which we know that two of them are very similar and that the third one behaves differently from the other two. A plot of their mean vectors can be seen in Fig.\ref{fig:1d}.
The first one models a Michalewictz function (defines as $f_1(x)=-\sin{x}\left(\sin{\frac{x^2}{\pi}}\right)^{2m}$) with parameter $m=50$, the second one a Michalewictz function with $m = 100$ and the third one models a parabola $x^2$.
%, where the benchmark function michalewictz is defined as $f_1(x)=-\sin{x}\left(\sin{\frac{x^2}{\pi}}\right)^{2m}$  \\
The correlation between the predicted means of the two Michalwicz GPs is $0.97$, and the average relative distance between them is $0.02$. In total, the distance calculated by our measure is $0.02$ over $1$, i.e., these GPs are very similar according to our function. 
On the other hand, when comparing the Michalewictz GP that has $m=100$ with the parabola, we obtain a correlation of $0.12$, which reflects how different their growth behaviour is, and an average relative distance of $0.27$. In total a distance of $0.72$ over 1.\\\\
We now compare three 2D bowl-shaped processes: one of them modelling a Styblinski-Tang function, another one an ellipsoid and the third one a sphere. \\The Syblinski-Tang function is given by $f_2(\textbf{x}) = \frac{1}{2}\sum_{i=1}^2\left(x_i^4-16x_i^2+5x_i\right)$, the ellipsoid is
$f_3(\textbf{x}) = \sum_{i=1}^2\sum_{j=1}^ix_j^2$ and the sphere function is $f_4(\textbf{x})= x_1^2+x_2^2$
Since the three are bowl shaped, they are somewhat similar, but from Fig.\ref{fig:bowl} we can tell that Styblinski-Tang is slightly different from the others. Indeed, when we compare the sphere and the ellipsoid processes we obtain a $0.94$ correlation between  their predicted mean vectors, and 0.06 is their average relative distance. Overall, the distance is $0.05$ which means these two functions are very similar. In contrast, when we compare the ellipsoid with the Styblinski-Tang process, we obtain a correlation of $0.74$: a value which is close to 1, reflecting the fact that both shapes are similar in a way, but not as close as in the previous comparison because the Styblinski-Tang bowl is different from the ellipsoid. The average relative distance is 0.13 (again a value which is greater than before) and the total distance of $0.22$ over 1, which shows that these processes are similar, but not as much as the previous ones.\\\\
We will now compare two processes modelling two functions that we know are not similar, illustrated in Fig.\ref{fig:misc}. The first one is a Griewank function $f_5(\textbf{x}) = \frac{1}{4000}\left(x_1^2+x_2^2\right)-\cos
(x_1)\cos\left(\frac{x_2}{\sqrt{2}}\right)+1$, and the second one a levy function $f_6(\textbf{x}) = \sin^2(\pi w_1) + (w_1-1)^2\left(1+10\sin^2(\pi w_1+1)\right)+ (w_2-1)^2\left(1+\sin^2(2\pi w_2)\right)$ where $w_i = 1 + (x_i-1)\frac{1}{4}$\\
Because of how different their shapes are, the correlation between the predicted mean is 0.04, almost no correlation at all. Their average relative distance is 0.16, and the distance between the processes is 0.75 over 1.\\\\
Finally, we will be comparing some Ackley functions. Recall that an ackley function depending on parameters $a$, $b$ and $c$ is given by 
$$f_7(\textbf{x)}=  -a\exp \left( -b\sqrt{\frac{1}{d}\sum_{i = 1}^2x_i^2}\right) 
-\exp \left( -\frac{1}{2}\sum_{i = 1}^2\cos\left(cx_i\right)\right) + a + e$$
We first compare two processes with fixed $a = 20$ and $b = 0.2$. The parameter $c$ equals $\pi$ for one process and $6\pi$ for the other, which results in unsimilar predicted means plots as can be seen in Fig. \ref{fig:ackley}. The correlation between them is 0.31, and the average relative distance 0.15, giving an overall distance of 0.55 over 1. Lastly, we compare two Ackley processes which, despite having different $a$ values, have a very similar shape (Fig.\ref{fig:ackley}). We chose $b=0.2, c=2\pi$ and $a=70$ for one process and $a=100$ for the other. Their correlation is very close to 1, 0.98, and their average relative distance is 0.01. A total distance of 0.01, which means these processes are indeed very similar in their shape.
\subsection{Many objective Bayesian Optimization algorithm experiments}
In this subsection, we describe a toy experiment, synthetic experiments and a real scenario involving the many objective scenario using our proposed objective reduction algorithm. We begin by presenting a toy experiment where the reduction of objectives is straightforward, although it must be discovered by the objective reduction algorithm. Then, we will follow the section by addressing two synthetic experiments with $4$ objectives involved. In all the experiments, we will perform a sensitivity analysis of the many objective Bayesian optimization algorithm that will show how this algorithm behaves according to its hyper-parameters. In order to measure the quality of the solution suggested by the algorithm we use the hyper-volume metric in all the experiments. We use the predictive entropy search for multi-objective Bayesian optimization (PESM) as acquisition function due to the fact that it provides state-of-the-art results solving multi-objective optimization problems \cite{hdezlobato16}. Both the similarity metric and the Many objective optimization algorithm have been coded on the Spearmint Bayesian optimization tool (\url{https://github.com/EduardoGarrido90/spearmint_ppesmoc/}). All the objectives are modelled by independent GPs using the Matérn covariance function. We use slice sampling to obtain $10$ samples of the hyper-parameter posterior distribution of the GPs to estimate the values of their hyperparameters. The PESM acquisition function is then obtained as the average of the acquisition functions computed by each of the hyperparameter samples. In our experiments, we compare the average of the predictive distributions obtained for each GP after the hyperparameter sampling process. Finally, in all the experiments that we perform in this subsection, we use the same similarity metric than in the previous subsection.

\subsubsection{Toy and synthetic experiments}
We begin with a toy example regarding the \textit{Branin} function, given by the following expression:
$$f_8(x,y)=\left( y - 5.1x^2/\left( 4\pi^2\right)  + 5x/\pi - 6 \right) ^2+ 10(1- 1/8\pi) \cos x + 10.$$
\begin{figure}
\includegraphics[width=\textwidth]{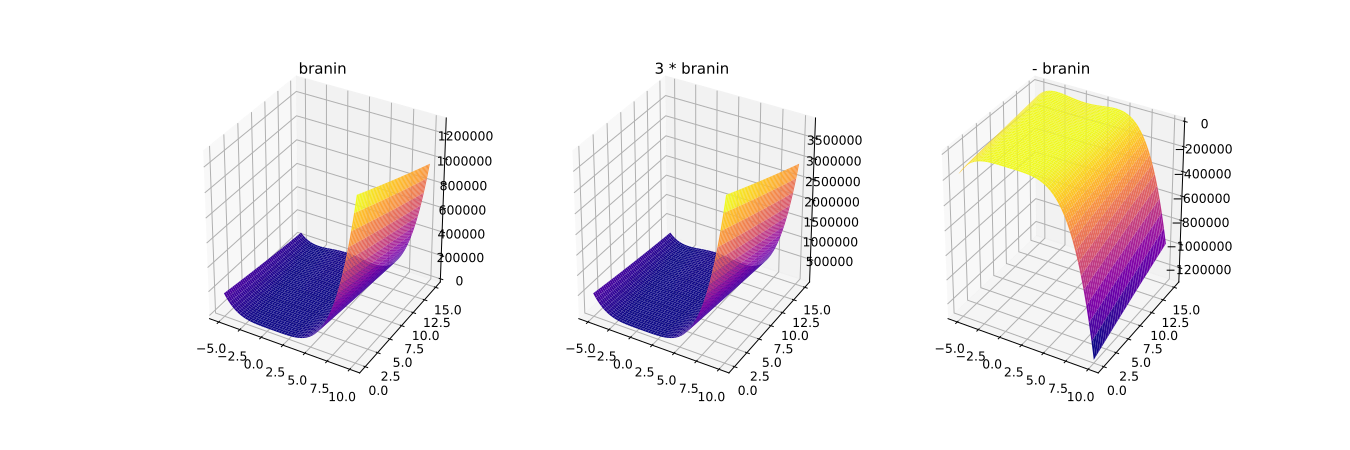}
\caption{Branin objective functions.}
\label{fig:exp2branin}
\end{figure}
In this toy experiment, we optimize the three objectives illustrated on Figure \ref{fig:exp2branin}: the first objective simply consist on a \textit{Branin} function, the second objective consist on another \textit{Branin} function triplicated and, finally, the third one is a \textit{Branin} function multiplied by $-1$. Clearly, the \textit{Branin} function and its triplicated version behave very similarly. Hence, we would like that our algorithm would be able to detect this similarity and delete one of them. In order to do so, we perform an experiment with $25$ iterations running our many objective Bayesian optimization algorithm. We perform a sensitivity analysis for the hyperparameter $\delta$, that applies the objective reduction at the $\delta$ iteration of the optimization. In particular, this hyperparameter has been tested for different values: $10$, $15$ and $20$. Finally, we also perform a sensitivity analysis concerning the hyperparameter $\varepsilon$, that represents whether two objectives are considered similar if their similarity metric regarding their predictive distributions is lower than $\varepsilon$. Concretely, it has been tested for the following values: $0.05$, $0.1$ and $0.2$. Hence, we use this toy problem to perform an exhaustive analysis of these hyperparameters and to study their influence on the objective reduction algorithm.

In this particular case, we can see that the first and second objectives are completely correlated. Concretely, we can see that their dependence relies on a linear function $f_2(\mathbf{x}) = 3 f_1(\mathbf{x})$. Hence, we expect that this dependence is found by the many objective Bayesian optimization algorithm. The results of the experiment are satisfactory in the sense that, for every mentioned configuration of the hyperparameters of the algorithm, it finds this correlation and deletes one of the two objectives just in the first iteration, according to the $\delta$ hyperparameter, that is able to do it. In particular, it detects the redundant objective in the $10$, $15$ and $20$ iterations according to $\delta$. Having added empirical evidence that the algorithm works on a toy experiment and is able to detect correlated objectives that supports the claim that our algorithm is useful in practice, we now focus our attention on the quality of the solution according to the hypervolume metric given by the Bayesian optimization method after deleting an objective. In particular, we would like to test the hypothesis that deleting correlated objectives do not influence heavily the quality of the final recommendation of the Bayesian optimization algorithm. In order to test this hypothesis, we run the \textit{Branin} experiment without applying the objective reduction algorithm and retrieve the hypervolume metric, that turns out to be $1143516861$. Let $h(\mathbf{f}(\mathbf{x}))$ be the hypervolume obtained by the final recommendation of the Bayesian optimization algorithm and $h_r(\mathbf{f}(\mathbf{x}))$ the hypervolume obtained by the recommendation applying the objective reduction algorithm. We expect that, after applying the objective reduction algorithm, $|h(\mathbf{f}(\mathbf{x}))-h_r(\mathbf{f}(\mathbf{x}))|$ is as lowest as possible. Having computed the hypervolume with all the objectives, we now apply objective reduction with different values of $\delta$, obtaining the results showed on Table \ref{TB:reduccion1}.
\begin{table}
\begin{center}
\begin{tabular}{||c | c c c||} 
 \hline
 Hypervolume & $\delta = 10$ & $\delta = 15$ & $\delta = 20$ \\ [0.5ex] 
 \hline%hline
 $\varepsilon = 0.05 $ & 1142795726 & 1142795726 & 1143013872 \\ 
 $\varepsilon = 0.10 $ & 1142682063 & 1142795726 & 1143013872 \\
 $\varepsilon = 0.20 $ & 1142682063 & 1142795726 & 1143013872\\ [1ex] 
\hline
\end{tabular}
\caption{Hypervolumes generated by the recommendations of the many objective Bayesian optimization on the \textit{Branin} experiment for different $\delta$ and $\varepsilon$ values.}
\end{center}
\label{fig:hypervolumes}
 \end{table}
 We can observe in Table 1, in all the cases, that the recommendation suggested by the many objective Bayesian optimization algorithm has a hypervolume that is close to the solution. Concretely, the worse case suggest a solution that is only a $0.07$ worse than the solution delivered by Bayesian optimization without reducing objectives. Moreover, and as we expect, we observe that the recommended solution is better if objective reduction is activated more lately in the optimization process. In other words, if $\delta$ is higher. This makes sense, as if we execute more iterations without reducing the objectives, we can expect that the GP predictive distribution match the objective function better, and as a consequence, we can expect that the similarity metric can better predict which objectives are correlated. Hence, we have added empirical evidence that support the claim that deleting an objective does not significantly hurt the performance of the Bayesian optimization algorithm but is able to save computational or other kind of resources in the optimization process. We also seen how the metric is able to successfully determine if the GP predictive distributions are correlated in practice. Another critical observation of the experiment is that the quality of the suggestions explored by the algorithm given by the maximization of the acquisition function $\alpha(\mathbf{x})$ in the first iterations after deleting an objective were lower as $\varepsilon$ is higher. A possible explanation of this phenomenon is that as $\varepsilon$ is higher, our reduction is performed more arbitrarily, that is, the GPs need to be less similar in order to be deleted from the optimization problem. Hence, exploration of new areas of the input space may be hurt according to a high $\varepsilon$ value. However, after these first iterations the results turned out to be good once again, possibly because the acquisition function $\alpha(\mathbf{x})$ explores again after exploiting promising results according to the new objectives. 
 
\begin{figure}
\includegraphics[width=\textwidth]{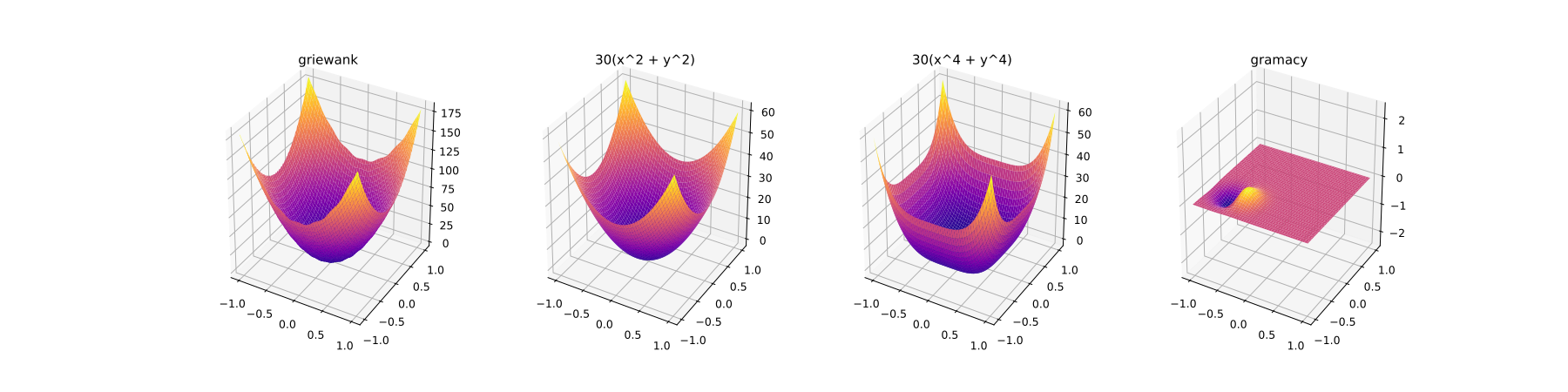}
\caption{Objectives based on paraboloids and other functions.}
\label{fig:expbowls}
\end{figure}

 In order to test the efficacy of the algorithm when a higher number of objectives are involved, we test its performance in a $4$ objective problem dealing with the following objective functions. Concretely, the first is a \textit{Griewank} function, given by the expression $1 -\cos (x) \cos(y / \sqrt{2}) + (x^2 + y^2)/4000 $, the second black-box is a two grade paraboloid given by ($x^2 + y^2$), the third function is another paraboloid of grade four, whose analytical expression is ($x^4 + y^4$) and, finally, the fourth black-box is a \textit{Gramacy} function that corresponds to the following expression: $x\mathrm{e} ^{-x^2 -y^2}$. The four objectives are illustrated on Figure \ref{fig:expbowls}. As it can be seen on that figure, the 
 \textit{Griewank} function is similar to a paraboloid, but its surface is full of local optima, making it different from a paraboloid in a wide range of points but similar in the whole range of the input space. We would also like to test whether the algorithm is able to reduce more than $1$ objective in the whole optimization process.
In this case, we have set the maximum number of iterations of the optimization to $30$ and we have carried out the experiment with two combinations of hyperparameters: the first with a looser reduction ($\delta = 15, \varepsilon = 0.10 $) and the second time an stricter reduction criterion ($ \delta = 20, \varepsilon = 0.05 $). Most critically, the hypervolume of the problem solution in this case, without applying the reduction algorithm, was $1000398078127$.

This time, in the first execution of the many objective Bayesian optimization algorithm ($ \delta = 15, \varepsilon = 0.10 $), the GP corresponding to the paraboloid of degree two was eliminated in the sixteenth iteration because it was very similar to the \textit{Griewank} objective function. Most importantly, the other paraboloid was eliminated in the twenty-first iteration because it is also considered equal to the \textit{Griewank} function. In this case, the recommended solution has hypervolume $1000397940485$ (which results in an error of $ 1.37 \times 10 ^ {- 7} $) and it is scanned between the first and second reduction. The image clearly shows why the \textit{Griewank} function and the paraboloid of degree two are considered equal. Although the other paraboloid is not so similar to these two functions, as a result of considering a high value of $\varepsilon=0.10$, this objective function is also removed. The function that indisputably does not resemble the others is the \textit{Gramacy} function, and therefore, this objective is not reduced by the many objective Bayesian optimization algorithm giving empirical evidence that the similarity metric correctly identifies correlated and independent objectives.

In the second execution, regarding the hyperparameters $\delta = 20, \varepsilon = 0.05 $, the GP corresponding to the paraboloid of degree two is removed after the twentieth iteration (the first in which targets can be reduced). No other targets are removed before the 30th and final iteration. The recommended solution in this case has a hypervolume $1000397851175$, with a corresponding error of $2.26 \times 10 ^{-7}$ and is scanned after the first shrink. In this case, delaying the reduction of targets and reducing the value of $\epsilon $ to $0.05$ causes only one target to be reduced and the paraboloid of degree four is not considered sufficiently similar to the other paraboloids. We can hence see how altering the value of the hyperparameters involve that the algorithm behaves differently, showing how the practitioner can adjust the similarity metric depending on its needs, basically a tradeoff between adquiring a high quality recommendation versus saving resources.

\begin{figure}
\includegraphics[width=\textwidth]{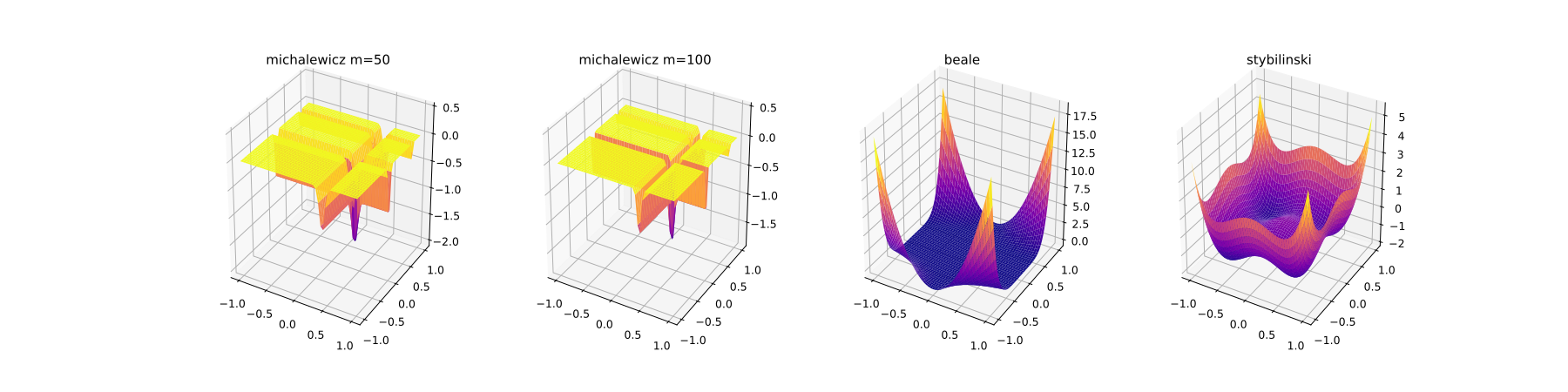}
\caption{Michalewicz objective functions.}
\label{fig:exp2mich}
\end{figure}

After having tested that the algorithm is able to reduce more than objective without incurring in a huge loss of quality of the final recommendation and having seen that varying its hyperparameters incurs in a different behaviour of the algorithm we would like to continue our analysis comparing different correlated shapes of objective functions to test whether the objective reduction optimization algorithm is able to identify those correlations. In particular, in this experiment, we have optimized the four functions illustrated on Figure \ref{fig:exp2mich}. Concretely, two \textit{Michalewicz} functions given by the expression 
$f_9(x,y) = -\sin x\sin^{2m}\left(x^2/\pi\right) -\sin y\sin^{2m}\left(2y^2/\pi\right),$
varying $m$, a \textit{Beale} function corresponding to the following expression: $f_{10}(x,y) = \left(1.5-x-xy\right)^2 + \left(2.25-x-xy^2\right)^2 + \left(2.625-x-xy* 3\right)^2$ and a \textit{Stybilinski-Tang} black-box objective function.
In this experiment, we have set the maximum number of iterations to $30$ and we have performed two tests. The first has been carried out with two \textit{Michalewicz} functions, one with $ m = 75 $ and the other with $ m = 100 $, and with the hyperparameter values $ \ delta = 10, \ varepsilon = 20 $. For this case, the two \textit{Michalewicz} functions were very similar and one of them was removed in the 21st iteration. The solution to the problem had hypervolume $1004234499054$ and the algorithm gives a recommendation with hypervolume $1003947732936$ (that basically consists on an $0.02 \%$ error). In the second test, we have used a \textit{Michalewicz} function with $m = 50$ and another with $m = 100$. Since these functions are more different from each other than those in the first test, we have reduced $\varepsilon$ to $0.05$. With this, we have demanded that the algorithm be more certain of the similarity of two targets before eliminating one of them. In this case, one of the two \textit{Michalewicz} functions has been deleted in the 27th iteration. The hypervolume of the solution of the problem was $1004348965464$ and that of the algorithm recommendation was $10043488477896$ (giving an error of just $4.85 \times 10^{-7}$). Most critically, this is the only test that we have executed where the solution is explored prior to target reduction, which is the ideal case scenario, involving in a reduction of costs without incurring in a poorer quality solution.

\subsubsection{Real Experiment}
In the previous experiments, we have given lots of empirical evidence supporting the claims that our proposed algorithm efficiently reduces objective functions when their predictive distributions are correlated and not losing quality of their recommendation. We believe that the contribution and benefits of the algorithm are clear with the described examples. Now, we provide a real case scenario where we could apply our proposed method to show its effectiveness in practice. In particular, we applied our objective reduction algorithm to the problem of the hyper-parameter tuning of a deep neural network that classifies images in digits between 0 and 9. For this purpose, we have used the Digits dataset, \href{https://scikit-learn.org/stable/auto_examples/datasets/plot_digits_last_image.html}, a small dataset of $1797$ images containing handwritten digits. In this experiment, we have considered an input space given by the number of hidden layers of the deep neural network and the number of neurons in each hidden layer. Concerning the objectives of the optimization problem, we are going to simultaneously minimize the prediction time, the network size and the prediction error of the deep neural networks. We measure the prediction time of the deep neural network by predicting a batch of unlabelled instances. The network size of the deep neural network is measured by the size of the file where it is stored. We hypothesize that the prediction time of the deep neural network and the size of the file where it is stored are correlated. Neural networks with lots of layers and neurons will demand more space to be stored and will spend more time to predict a batch of instances. However, minimizing those objectives will, theoretically, incur in a lower estimation of the generalization error in huge datasets. Hence, we expect that our algorithm would be able to delete the prediction time or network size objectives. As a consequence, we do not really care about an accurate estimation of the generalization error but in detecting if the algorithm is able to detect the correlation between the prediction time or network size objectives, so, for this experiment, we allow the range of the hidden layers and number of neurons to be really huge and we simply do not train the weights of the deep neural network, leaving them at random. By doing it so, we can test different configurations of huge deep neural networks that could be applied for current real image classification problems and test whether our algorithm works without having to train the deep neural networks, that is unfeasible giving our resources and it really does not add any value to test the hypothesis that this particular experiment intends to solve. Concretely, detecting correlated objectives in a real case scenario.

The input space of the experiment consists on varying the number of hidden layers in the following interval: $[3, 100]$ and the number of neurons in each layer in the interval: $[3, 300]$. We believe that the objectives of prediction time and size of the neural network may be very correlated so we expect that the many objective Bayesian optimization algorithm will discover this correlation very fast. In order to test this hypothesis, we set the maximum number of iterations to $40$. 

In the first test, we fix the hyperparameters to $\delta = 15$ y $\varepsilon = 0.1$. In this particular scenario, the many objective Bayesian optimization algorithm detect that the prediction time is similar to the size of the network at the $16$ iteration. In particular, the distance obtained by the similarity metric between the predictive distributions of the GPs that model those objectives is just $0.0006$. However, the distance between those objectives and the predictive distribution of the prediction error is much higher, as expected, concretely $0.75$. As a consequence, the many objective Bayesian optimization algorithm deleted the objective representing the size of the network. 

We finally run another test of our algorithm using different hyperparameters, concretely $\delta = 20$, $\varepsilon = 0.05$. Our purpose with this final scenario is just to test whether applying the reduction after several iterations more, when the predictive distribution of the GPs resemble more the objective functions, and being more strict with the $\varepsilon$ hyperparameter, would also incur in a deletion of a redundant objective. The results that we obtained were also satisfying. In particular, our algorithm detected a redundant objective at the $21$ iteration. Moreover, the similarity metric computed that the distance between the predictive distributions of the prediction time and the size of the network was just $0.0004$, even less that in the previous experiment, and the other distances were higher, $0.7502$. As a consequence, the GP that models the size of the network was also selected as redundant in this experiment and deleted, as we expected. 

\section{Conclusions}
In this paper, we have proposed a similarity metric for GP predictive distributions in order to be used in a many objective BO scenario \cite{ishibuchi2008evolutionary}. Thanks to this metric, we are able to measure how similar are the predictions given by two GP predictive distributions. We have illustrated the results of this measure in a set of synthetic and benchmark functions. Having given empirical evidence of its usefulness, this measure can be applied for multi-objective BO, that could be parallel or constrained, when there are $3$ or more objectives.

We use this similarity metric in a many objective BO algorithm to detect similar objectives in terms of the predictive distributions of the GPs that model them. In particular, the algorithm reduces the number of the objectives of the problem if it detects that they are similar according to the similarity metric. Hence, it converts a many objective BO setting in a multi-objective BO setting. Concretely, the reduction is only applied if the objectives are similar, not hurting severely the quality of the final recommendation of the BO algorithm. We present toy, synthetic and a real experiment concerning hyper-parameter tuning of deep neural network where we present empirical evidence of the effectiveness of our algorithm. In particular, it is able to detect redundant objectives, the reduction of the objectives does not hurt the final recommendation and we can configure the metric depending on how important does the practitioner consider to be to save resources, that is detecting similar objectives, and the quality of the recommendation. 

We consider that our contribution is very valuable for the BO community as, as far as we know, this is the first algorithm that targets the many objective Bayesian optimization setting. As further work, we can also incorporate the uncertainty of the GP prediction mean in the similarity metric. The intuition is that the areas whose uncertainty on the prediction is low need to add more weight that unexplored areas. Suppose that two GP predictive distributions differ in their predictions in an unexplored area. It is fair to assume that they may be more similar than if they differ on an explored area. We intend to incorporate this property to the GP similarity by measuring the absolute error of the Frobenius norm of the GP covariance matrices. We would also like to test this metric in a parallel constrained many-objective setting where the problem does also consider a set of constraints that could also be correlated \cite{garrido2020parallel}. In particular, deleting constraints may discover a feasible configuration more rapidly, although with a higher risk. 

\section*{Acknowledgments}

Authors gratefully acknowledge the use of the facilities of Centro de Computacion Cientifica (CCC)
at Universidad Autónoma de Madrid. The authors also acknowledge financial support from Spanish Plan Nacional I+D+i, grants TIN2016-76406-P and from PID2019-106827GB-I00 /AEI /10.13039/501100011033.

\bibliographystyle{plain}
\bibliography{main}
\end{document}